\documentclass[runningheads]{llncs}
\usepackage{graphicx}

\usepackage{tikz}
\usepackage{comment}
\usepackage{amsmath,amssymb} 
\usepackage{color}

\usepackage{algorithm}
\usepackage{algorithmic}

\usepackage{setspace}

\usepackage{subfigure}
\usepackage{multirow}
\usepackage{hhline}
\usepackage{enumitem}
\usepackage[T1]{fontenc}
\usepackage{caption}

\begin{document}
\pagestyle{headings}
\mainmatter
\def\ECCVSubNumber{6101}  

\title{Learning to Localize Actions from Moments\thanks{{\small This work was performed at JD AI Research.}}} 

\titlerunning{Learning to Localize Actions from Moments}
%
\author{Fuchen Long\inst{1} \and
Ting Yao\inst{2} \and
Zhaofan Qiu\inst{1} \and
Xinmei Tian\inst{1} \and \\
Jiebo Luo\inst{3} \and
Tao Mei\inst{2}}
\authorrunning{F. Long, T. Yao, Z. Qiu, X. Tian, J. Luo and T. Mei.}
%
\institute{University of Science and Technology of China, Hefei, China \and
JD AI Research, Beijing, China \and
University of Rochester, Rochester, NY USA\\
\email{\{longfc.ustc, tingyao.ustc, zhaofanqiu\}@gmail.com; xinmei@ustc.edu.cn; \\
jluo@cs.rochester.edu; tmei@jd.com}}
\maketitle

\begin{abstract}
  With the knowledge of action moments (i.e., trimmed video clips that each contains an action instance), humans could routinely localize an action temporally in an untrimmed video. Nevertheless, most practical methods still require all training videos to be labeled with temporal annotations (action category and temporal boundary) and develop the models in a fully-supervised manner, despite expensive labeling efforts and inapplicable to new categories. In this paper, we introduce a new design of transfer learning type to learn action localization for a large set of action categories, but only on action moments from the categories of interest and temporal annotations of untrimmed videos from a small set of action classes. Specifically, we present Action Herald Networks (AherNet) that integrate such design into an one-stage action localization framework. Technically, a weight transfer function is uniquely devised to build the transformation between classification of action moments or foreground video segments and action localization in synthetic contextual moments or untrimmed videos. The context of each moment is learnt through the adversarial mechanism to differentiate the generated features from those of background in untrimmed videos. Extensive experiments are conducted on the learning both across the splits of ActivityNet v1.3 and from THUMOS14 to ActivityNet v1.3. Our AherNet demonstrates the superiority even comparing to most fully-supervised action localization methods. More remarkably, we train AherNet to localize actions from 600 categories on the leverage of action moments in Kinetics-600 and temporal annotations from 200 classes in ActivityNet~v1.3. Source code and data are available at \url{https://github.com/FuchenUSTC/AherNet}.
\end{abstract}

\section{Introduction}
With the tremendous increase in Internet bandwidth and the power of the cloud, video data is growing explosively and video-based intelligent services are becoming gradually accessible to ordinary users. This trend encourages the development of recent technological advances, which facilitates a variety of video understanding applications \cite{Buch:CVPR17,Dong:ECCV18,Dong:MM19,Lu:TMM14}. In between, one of the most fundamental challenges is the process of temporal action localization \cite{Gaidon:PAMI13,Geest:ECCV16,Lea:CVPR17,Lin:ECCV18,Shou:CVPR16,Xiong:ICCV17}, which is to predict the temporal boundary of each action in an untrimmed video and categorize each action according to visual content as well. Most existing action localization systems still perform ``intensive manual labeling'' to collect temporal annotations (action category and temporal boundary) of actions in untrimmed videos and then train localization models in a fully-supervised manner. Such paradigm requires strong supervision, which is expensive to annotate for new categories and thus limits the number of action categories. In the meantime, there are various datasets (e.g., Kinetics \cite{Kinetics:600}) which include expert labeled data of trimmed action moments for action recognition. A valid question then emerges as is it possible to achieve action localization for a large set of categories, with only trimmed action moments from these categories and temporal annotations from a small set of action classes? If possible, it is readily to adapt state-of-the-art action localization methods to support thousands of action categories in real-world deployment.

\begin{figure}[!tb]
         \centering\includegraphics[width=1.0\textwidth]{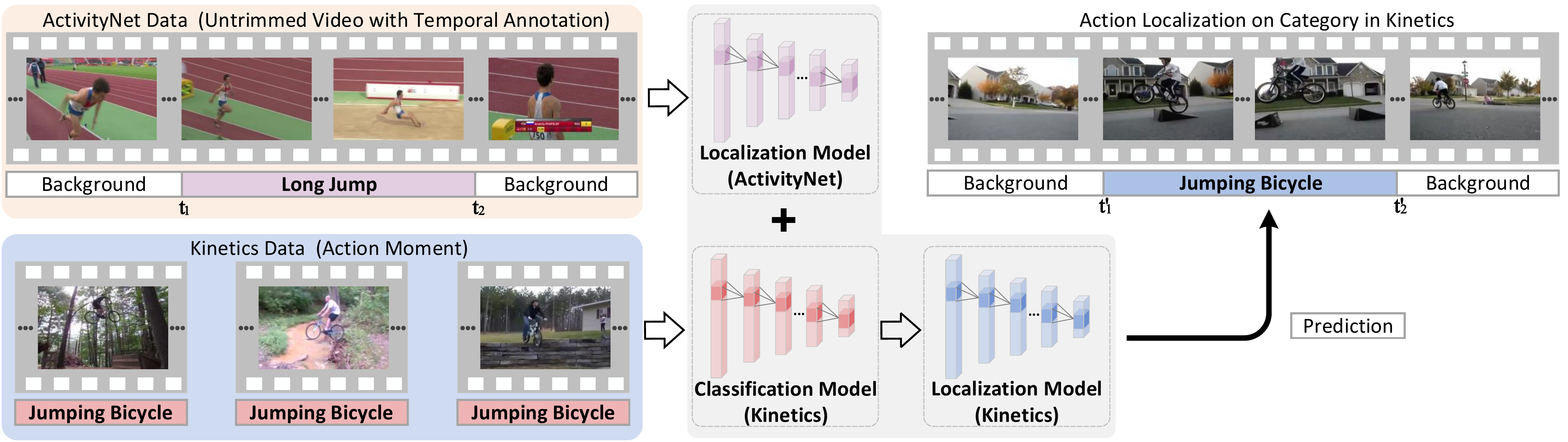}
         \vspace{-0.28in}
         \caption{Action localization modeling for a large set of categories based on only action moments of these categories (e.g., Kinetics \cite{Kinetics:600}) and untrimmed videos from a small set of categories with temporal annotations (e.g., ActivityNet \cite{ActivityNet}).}
         \vspace{-0.30in}
         \label{fig1:1}
\end{figure}

With this motivation, Figure \ref{fig1:1} conceptually depicts the pipeline of action localization in our work. Given a large set of categories which have only trimmed action moments (e.g., Kinetics \cite{Kinetics:600}) and a small set of classes which have fully temporal annotations on untrimmed videos (e.g., ActivityNet \cite{ActivityNet}), we aim for a model that enables to temporally localize and recognize actions from the large set of categories. Note that the categories in the two sets could be completely different. The main difficulties inherently originate from two aspects: 1) how to build the connection between classification and localization? 2) how to hallucinate the context or background of an action moment in training? We propose to mitigate the first issue through the design of weight transfer. In view that action localization generally consists of temporal action proposal and temporal action classification, the network weights for temporal action classification could be derived from those for action recognition of trimmed videos. In our case, the trimmed videos are either foreground video segments in untrimmed videos or action moments. As such, the weight transfer is considered as a bridge between classification and localization. We utilize the recipe of adversarial learning to alleviate the second issue. A discriminator is devised to differentiate the generated context features from those of background in untrimmed videos.

By consolidating the idea of learning action localization models on a mixture of action moments and fully temporal annotations, we present a new Action Herald Networks (AherNet) in an one-stage localization framework. AherNet mainly includes two modules, i.e., weight transfer between classification and localization on untrimmed videos with temporal annotations, and localization modeling on action moments with synthetic contexts. On one hand, the first module naturally constructs a correspondence between action localization in an untrimmed video and action classification of ``action moment'', i.e., the foreground video segment extracted from the untrimmed video. Technically, we learn a weight transfer function which transforms network parameters for foreground segment classification to those for temporal action classification in localization on untrimmed videos. On the other hand, to simulate action localization on action moments data, we hallucinate the features of context or background of an action moment via adversarial learning. The connection between action moment classification and localization of the action from the context is also built by the weight transfer function, whose parameters are shared. The whole AherNet is end-to-end optimized by minimizing proposal loss, classification loss and adversarial loss.

The main contribution of this work is a new paradigm between supervised and weakly-supervised training, that enables action localization models to support thousands of action categories, with only trimmed action moments from these categories and temporal annotations from a small set of classes.
This also leads to the elegant view of how to bridge the task of classification and localization, and how to produce the context of action moments to simulate localization in training, which are problems not yet fully understood.

\section{Related Work}

\textbf{Temporal Action Localization.}
We briefly group the temporal action localization into two categories: two-stage and one-stage action localization. Two-stage action localization approaches \cite{Heilbron:CVPR17,Shou:CVPR17,Shou:ECCV18B,Xu:ICCV17,Zeng:ICCV19,Xiong:ICCV17} first detect temporal action proposals \cite{Buch:CVPR17,Escorcia:ECCV16,Gao:ECCV18,Gao:ICCV17,Lin:ECCV18,Long:TMM20,Oneata:ICCV13,Tang:ICCV13,Yuan:CVPR16} and then classify \cite{Qiu:ICCV17,Qiu:CVPR19} the proposals into known action classes. For instance, Buch \emph{et al.} \cite{Buch:CVPR17} develop a recurrent GRU-based action proposal model followed by a S-CNN \cite{Shou:CVPR16} classifier for localization.
To further facilitate action localization by uniting separate optimization of two stages, there have been several one-stage techniques \cite{Buch:BMVC17,Chao:CVPR18,Lin:MM17,Long:CVPR19,Yeung:CVPR16} being proposed.
All these methods require the training data with fully temporal annotations. Instead, our AherNet models action localization for a large set of categories based on only action moments of these categories and untrimmed videos from a small set of categories with temporal annotations.

\textbf{Parameter Prediction.}
Parameter prediction in neural networks is capable of building the connections between the related tasks.
Several weight adaptation methods \cite{Hoffman:NIPS14,Kuen:ICCV19,Tang:CVPR16} learn specific matrix to adapt the image classification weights for object detection.
Most recently, Hu \emph{et al.} \cite{Hu:CVPR18} explore the direction of parameter transferring from object detection to instance segmentation by a general function, which enables the transformed Mask R-CNN \cite{He:ICCV17} to segment 3000 visual concepts.
In our work, we utilize the parameter prediction to bridge the task of classification and localization.

\textbf{Adversarial Learning.}
Inspired by the Generative Adversarial Networks (GAN) \cite{Goodfellow:NIPS14}, the adversarial learning has been widely used in various vision tasks, e.g., image translation \cite{Isola:CVPR17} and domain adaptation \cite{Ganin:ICML15,Tzeng:CVPR17}. The training processing of GAN \cite{Goodfellow:NIPS14} corresponds to a minimax two-player game to make the distribution of fake data close to the real data distribution. In the context of our work, we simulate action localization on action moments with generated action contexts. Through adversarial learning, the generated contextual features become indiscriminative from real background features of untrimmed video.

\textbf{Weakly-supervised Action Localization.}
The weakly-supervised action localization approaches \cite{Liu:CVPR19,Nguyen:CVPR18,Nguyen:ICCV19,Shi:CVPR20,Shou:ECCV18,Wang:CVPR17} only utilize the category supervision of untrimmed videos for localization, whose setting and scenario are different from our paradigm.
Most of them build an attention mechanism to detect actions.

In short, our work mainly focuses on a new learning paradigm of scaling action localization to a large set of categories. The proposal of AherNet contributes by studying not only bridging action classification and localization through weight transfer, but also how the generated context of action moments should be better leveraged to support action localization learning.

\begin{figure*}[!tb]
\centering\includegraphics[width=1.0\textwidth]{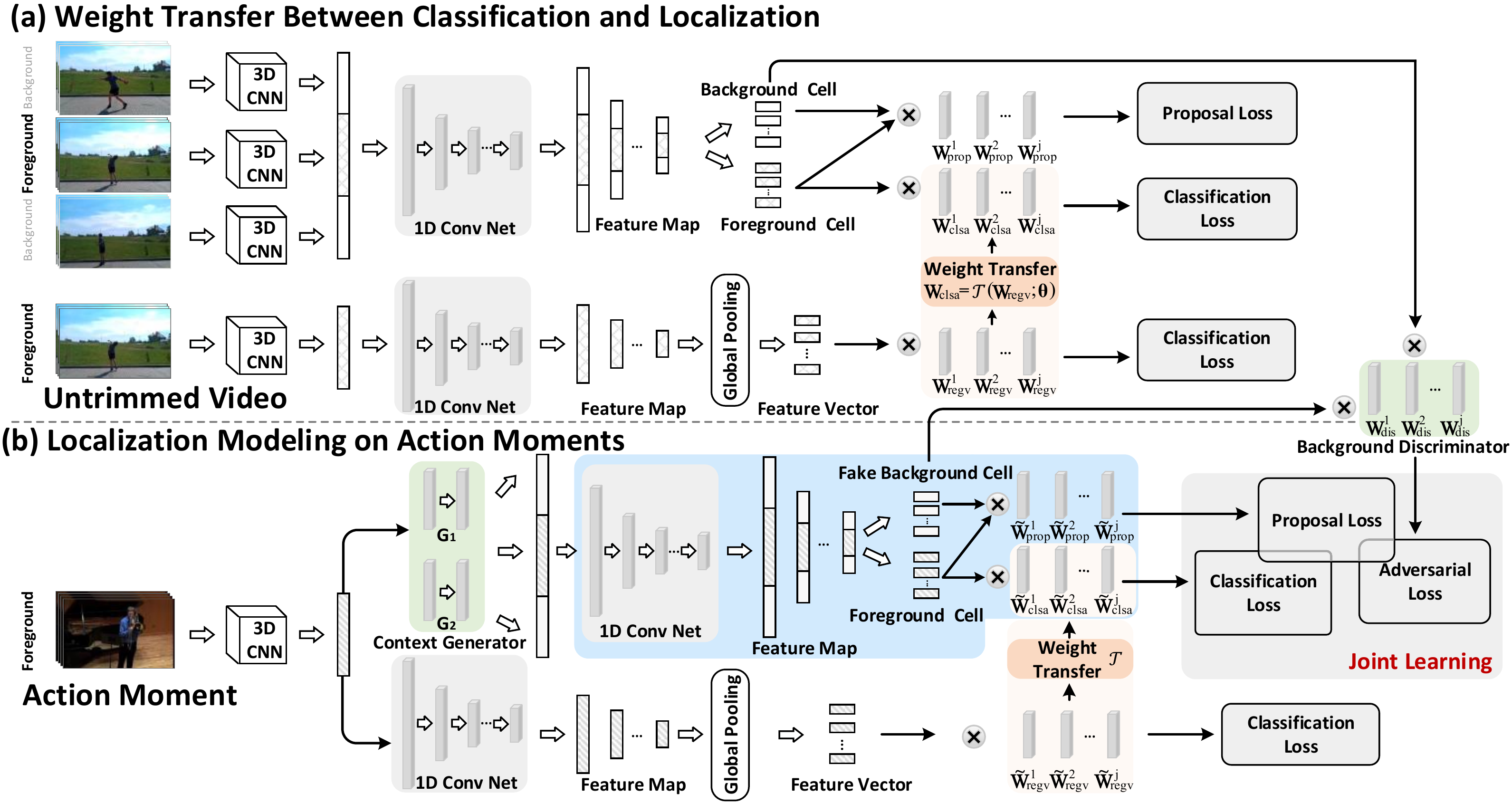}
\caption{An overview of our Action Herald Networks (AherNet) architecture. The foreground segments of untrimmed videos are first extracted as ``action moments.'' The input untrimmed video and foreground segment is encoded into a series of clip-level features via a 3D ConvNet, which are sequentially concatenated as a feature map, respectively. A cascaded of 1D convolutional layers is applied to generate multiple feature maps on different scales. For classification of foreground segment, global pooling is exploited on all cells of feature map to produce the features on each scale, which are projected via a matrix for segment-level classification. The matrix is adapted by a weight transfer function (orange box) to that used in action localization for untrimmed video. In localization, the adapted matrix is performed on each cell in the feature map to obtain the projection for temporal action classification. Similar process are implemented on action moments and the extensions with generated context. The synthetic contexts of moments are confused with the background of untrimmed videos via adversarial learning (green box) and the parameters of weight transfer function are shared. Our AherNet is jointly optimized with proposal loss, classification loss and adversarial loss. In the inference stage, only the localization part (blue box) learnt on the moments with contexts is utilized to predict action instances.}
\label{fig2:1}
\end{figure*}

\section{Action Herald Networks}
In this section we present the proposed Action Herald Networks (AherNet) in detail. Figure \ref{fig2:1} illustrates an overview of our architecture. It consists of two modules, i.e., weight transfer between classification and localization, and localization modeling on action moments.
Given an untrimmed video, the foreground video segment is extracted as the ``action moment.'' A 3D ConvNet is exploited as the base network to extract a sequence of clip-level features for the untrimmed video and foreground segment, respectively. Each feature sequence is concatenated into a feature map, followed by a cascaded of 1D temporal convolutional layers to output feature maps on different scales. For action classification of foreground segment, global pooling is employed on the features of all the cells in each feature map to produce the features on each scale, which are projected via a matrix for segment-level classification. Such matrix is adapted by a weight transfer function to that used in action localization for the untrimmed video. In that case, we perform the adapted matrix on each feature map to obtain the projection of the features of every cell (anchor) in that map for temporal action classification. Similar processes are implemented on action moments and the extensions with contexts. The features of contexts are hallucinated through adversarial learning and the parameters of the weight transfer function are shared.
The network is jointly optimized with proposal loss, classification loss and adversarial loss.

\subsection{Base Backbone}
We build our action localization model on a weight-sharing 1D convolutional networks.
Given an input untrimmed video or action moment, a sequence of clip-level features are extracted from a 3D ConvNet.
We concatenate all the features into one feature map and then feed the map into a cascaded of 1D convolutional layers (anchor layers) to generate multiple feature maps on eight temporal scales.
These feature maps are further exploited for action classification of the action moment or temporal action localization of the untrimmed video.

\subsection{Weight Transfer Between Classification and Localization}
Given the feature maps of an untrimmed video in 1D ConvNet, temporal boundary regression and action classification can be optimized for each anchor in the feature maps.
For an action moment or foreground segment, the representation of global pooling on each feature map is able to be used for segment-level classification.
In view that action localization task decomposes into temporal action proposal and classification, the parameters of temporal action classification in localization to predict the score of a specific action category could be derived from the weights of moments recognition for the same category.
To build the connection between the two tasks, we extract the foreground segment of untrimmed video as moment and learn a generic weight transfer function to transform parameters for foreground segment classification to those for temporal action classification in localization.

Specifically, in $j$-th feature map of foreground segment, global pooling is first employed on that map to produce a feature vector.
Then a matrix $\mathbf{W}^{j}_{regv,c}$ is utilized to project the feature vector into the probability of category $c$ for segment-level classification. As for localization on untrimmed video, we adopt a 1D convolutional layer with stride of 1 to obtain the score of each cell (anchor) in that map for anchor-level classification. The parameters in that 1D convolutional layer to predict score of category $c$ are denoted as $\mathbf{W}^{j}_{clsa,c}$. To bridge classification and localization for the specific category $c$, a generic weight transfer function $\mathcal{T}$ is introduced to predict $\mathbf{W}^{j}_{clsa,c}$ from $\mathbf{W}^{j}_{regv,c}$ as follows:
\begin{equation}\label{Eq2:1}
\mathbf{W}^{j}_{clsa,c} = \mathcal{T}(\mathbf{W}^{j}_{regv,c};\mathbf{\theta}^j),
\end{equation}
where \mbox{$\mathbf{\theta}^j$} are the learnt parameters irrespective of action category.
$\mathcal{T}$ can be implemented with one or two fully-connected layers activated by different functions.
Through sharing \mbox{$\mathbf{\theta}^j$} with the transfer module in $j$-th anchor layer between classification and localization on moments, $\mathcal{T}$ is generalized to the categories of action moments.
The weights of segment-level classification for those categories can be transferred to the weights of anchor-level classification. As such, the weight transfer function is considered as a bridge to leverage the knowledge encoded in the action classification weights for action localization~learning.

\subsection{Localization Modeling on Action Moments}

\begin{figure}[!tb]
\centering\includegraphics[width=0.76\textwidth]{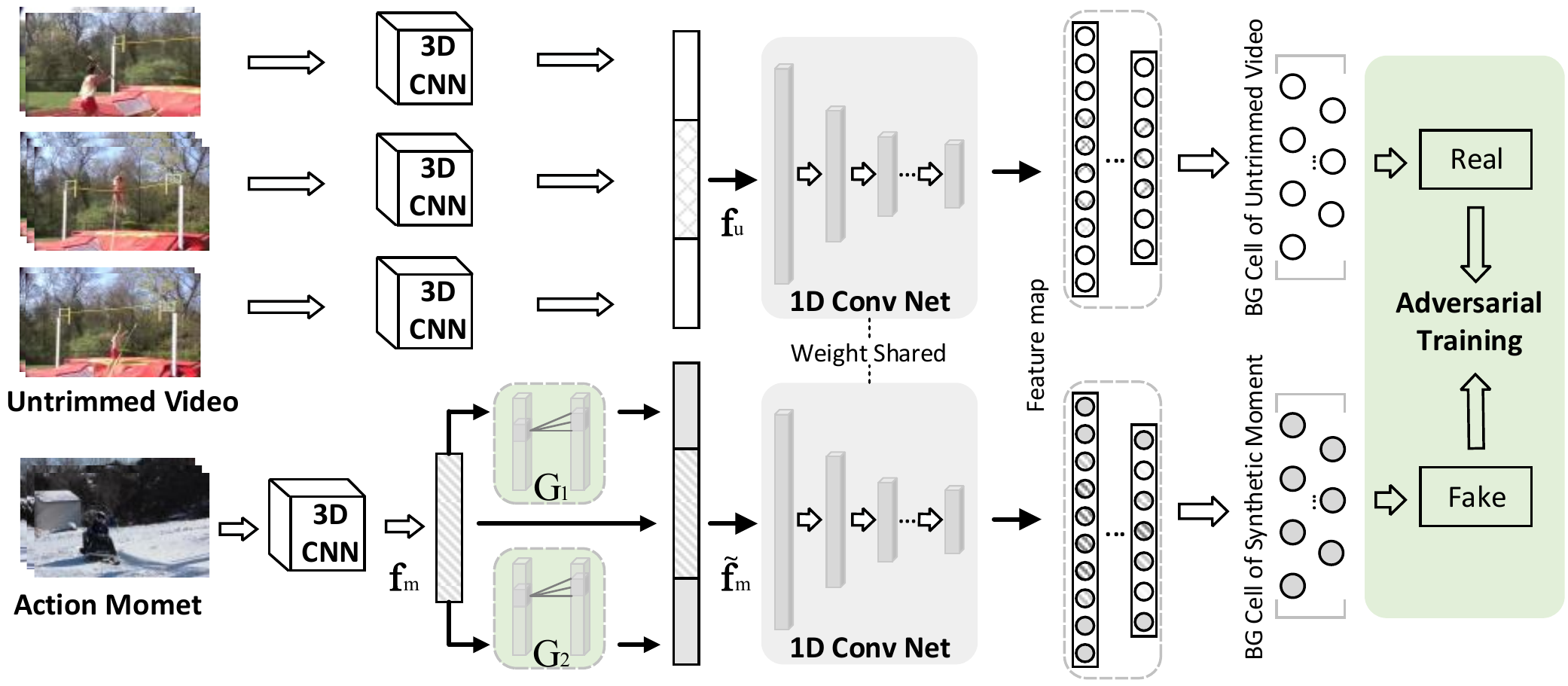}
\vspace{-0.10in}
\caption{Action context generation through adversarial learning. (BG: background)}
\vspace{-0.10in}
\label{fig2:2}
\end{figure}

With the obtained anchor-level classification weights predicted by weight transfer function on action moments, we still can not perform action localization training since there is no background for optimizing temporal action proposal.
To leverage action moments data for training localization model, a natural way is to hallucinate the background of moment to synthetize a complete action video. We therefore propose to generate action moment contextual features for localization modeling in an adversarial manner.

Figure \ref{fig2:2} illustrates the process of action context generation for action moments.
We denote the concatenated feature map of action moment and untrimmed video extracted by 3D ConvNet as $\mathbf{f}_m$ and $\mathbf{{f}}_u$.
Taking $\mathbf{f}_m$ as prior knowledge, two generators ($G_1$ and $G_2$) with the structure of two 1D convolutional layers are followed to synthesize the starting and ending contextual feature, respectively. The synthetic moments feature $\mathbf{\tilde{f}}_m$ is generated by concatenating $\mathbf{f}_m$ with the two generated contextual features as follows:
\begin{equation}\label{Eq2:2}
\mathbf{\tilde{f}}_m = A(G_1(\mathbf{f}_m),\mathbf{f}_m, G_2(\mathbf{f}_m)),
\end{equation}
where $A$ denotes the concatenation operation.
By feeding the synthetic feature $\mathbf{\tilde{f}}_m$ and the original feature $\mathbf{{f}}_u$ of untrimmed videos into the 1D convolutional networks of localization model, multiple feature maps are produced on different scales. Each cell (anchor) in the $j$-th feature map reflects an action proposal, and the default temporal boundary of the $t$-th cell is defined as:
\begin{equation}\label{Eq2:3}
m_c = (t+0.5)/T^{j},~~~~m_w = r_d/T^{j},
\end{equation}
where $m_c$ and $m_w$ are the center location and width. $T^j$ and $r_d$ represents the temporal length and scale ratio, respectively. For each cell, we denote the intersection over union (IoU) between the corresponding proposal and it's closest ground truth as $g_{iou}$. If the $g_{iou}$ is larger than $0.8$, we regard the cell as foreground cell. If the $g_{iou}$ is lower than $0.3$, it will be set as background cell. In each feature map, a discriminator is introduced to differentiate the background cells of synthetic moments from those of untrimmed videos.
The simulation of action localization is employed on the concatenated synthetic feature.

Through adversarial learning, the contextual features of synthetic moments tend to be real through the guidance from those of untrimmed videos.
Meanwhile, the anchor-level classification loss in localization modeling serves as a conditional constraint for adversarial training.
The loss alleviates the generation of trivial background features and regularizes the generated context of each moment to preserve semantic information of action category.

\begin{algorithm}[htb]
       \caption{AherNet Optimization}
       \label{OSA}
       \begin{algorithmic}[1]
              \REQUIRE ~~\\
              Localization model $\mathbf{M}$ pre-trained on untrimmed videos; \\ 
              Maximum number of iteration $N$; \\
              \ENSURE ~~\\
              Localization model $\mathbf{\tilde{M}}$ for action categories from moment set; \\
              \STATE Initialize the 1D ConvNet with $\mathbf{M}$, the iterative count $n = 1$;
              \FOR{n = 1 to N}
              \STATE Optimize $L_{reg}$ for foregrounds and moments to learn $\mathbf{W}_{regv}$ and $\mathbf{\tilde{W}}_{regv}$;
              \STATE Fix $\mathbf{W}_{regv}$, optimize $L_{cls}$ and $L_{prop}$ of untrimmed videos to learn $\theta$;
              \STATE Apply $\theta$ to $\mathbf{\tilde{W}}_{regv}$ and obtain $\mathbf{\tilde{W}}_{clsa}$ for synthetic moments classification;
              \STATE Fix 1D ConvNet, optimize context generators through ${L}_{\mathrm{ad}_G}$, $L_{cls}$ and $L_{prop}$ of synthetic moments. Then fix context generators, optimize 1D ConvNet through ${L}_{\mathrm{ad}_D}$, $L_{cls}$ and $L_{prop}$ of synthetic moments;
              \ENDFOR
              \RETURN $\mathbf{\tilde{M}}$
       \end{algorithmic}
\end{algorithm}

\subsection{Network Optimization}

Given the global pooling feature vector $f^j_p$ of $j$-th feature map, the segment-level classification loss ($L_{reg}$) for foreground segment or action moment is formulated via softmax loss:
\begin{equation}\label{Eq2:4}
L_{reg} = -\sum\limits_{n=0}^{C-1}I_{n=c}\log(p^{j}_{n}),
\end{equation}
where $C$ represents the total number of action categories in untrimmed video set or moment set. The indicator function $I_{n=c}=1$ if $n$ equals to ground truth label $c$, otherwise $I_{n=c}=0$. The probability $p^{j}_{n}$ is projected by $\mathbf{W}^{j}_{regv}$ on $f^j_p$.

For the optimization of action localization, three 1D-conv layers are utilized on each feature map of untrimmed video or synthetic moment to predict anchor-level classification scores, offset parameters and overlap parameter for~each cell (anchor).
The anchor-level classification scores are predicted by transformed weights $\mathbf{W}_{clsa}$ and the formulation of loss function $L_{cls}$ is the same with Eq.(\ref{Eq2:4}). The offset parameters $(\Delta c, \Delta w)$ denote temporal offsets relative to default center location $m_c$ and width $m_w$, which are leveraged to adjust temporal~coordinate:
\begin{equation}\label{Eq2:5}
\begin{split}
\varphi_c = m_c + \alpha_1 m_w \Delta c ~~~{\rm{and}} ~~~\varphi_w = m_w \exp{(\alpha_2 \Delta w)}~,
\end{split}
\end{equation}
where \mbox{$\varphi_c$}, \mbox{$\varphi_w$} are refined center location and width of the corresponding proposal. $\alpha _1$ and $\alpha _2$ are used to balance the impact of temporal offsets.
The offset loss is devised as Smooth L1 loss \cite{Girshick:ICCV15} (\mbox{$S_{L1}$}) between the foreground proposal and the closest ground truth, which is computed by
\begin{equation}\label{Eq2:6}
{L}_{of} = S_{L1}(\varphi_{c}-g_{c})+S_{L1}(\varphi_{w}-g_{w}),
\end{equation}
where \mbox{$g_{c}$} and \mbox{$g_{w}$} represents the center location and width of the proposal's closest ground truth instance, respectively.
Furthermore, we define an overlap parameter $y_{ov}$ to regress IoU between the proposal and it's closest ground truth for proposal re-ranking in localization. The mean square error (MSE) loss is adopted to optimize it as follows:
\begin{equation}\label{Eq2:7}
L_{ov} = (y_{ov} - g_{iou})^2.
\end{equation}
Since both of the offset loss ($L_{of}$) and overlap loss ($L_{ov}$) are optimized for temporal action proposal, the sum of the two is regarded as the proposal loss ($L_{prop}$).

In the moment context generation stage, we define $G$ as context generators of action moments, while $D$ represents the discriminator of background cell on the feature map. We denote $\mathcal{{F}}_u$ and $\mathcal{{F}}_m$ as the set of extracted feature maps of untrimmed video and moment set, respectively.
After producing the background cells $b_u$ and $b_m$ of each set, the adversarial loss is formulated as
\begin{equation}\label{Eq2:9}
\begin{split}
&{L}_{{ad}_D}  = - E_{\mathbf{f}_u \sim \mathcal{F}_u}[\log(D(b_u;\mathbf{f}_u))] - E_{\mathbf{f}_m \sim \mathcal{F}_m}[\log(1-D(b_m;G(\mathbf{f}_m)))],\\
&{L}_{{ad}_G}  = - E_{\mathbf{f}_m \sim \mathcal{F}_m}[\log(D(b_m;G(\mathbf{f}_m)))].
\end{split}
\end{equation}

The overall training objective of our AherNet is formulated as a multi-task loss by integrating classification loss in segment-level ($L_{reg}$) and anchor-level ($L_{cls}$), proposal loss ($L_{prop}$) and adversarial loss ($L_{ad}$). The weight-sharing 1D convolutional networks of localization model are first pre-trained on untrimmed videos for initialization.
Then we propose an alternating training strategy in each iteration to optimize the whole networks in an end-to-end manner.
Algorithm \ref{OSA} details the optimization strategy of our AherNet.

\subsection{Inference and Post-processing}
During prediction of action localization on action moment set, the context generators have been removed.
The final ranking score $s_f$ of each candidate action proposal is calculated by anchor-level classification scores $\mathbf{p}=[p_0, p_1, ..., p_{C-1}]$ and overlap parameter $y_{ov}$ with $s_f = \max(\mathbf{p}) \cdot y_{ov}$.
Given the predicted action instance $\phi=\{\varphi_{c},\varphi_{w},C_a,s_f\}$ with refined boundary ($\varphi_{c},\varphi_{w}$), predicted action label $C_a$, and ranking score $s_f$, we employ the non-maximum suppression (NMS) for post-processing.

\section{Experiments}
We empirically verify the merit of our AherNet by conducting the experiments of temporal action localization across three different settings with three popular video benchmarks: ActivityNet v1.3 \cite{ActivityNet}, THUMOS14 \cite{Thumos} and Kinetics-600~\cite{Kinetics:600}.

\subsection{Datasets}
The \textbf{ActivityNet v1.3} dataset contains 19,994 videos in 200 classes collected from YouTube. The dataset is divided into three disjoint subsets: training, validation and testing, by 2:1:1. All the videos in the dataset have temporal annotations. The labels of testing set are not publicly available and the performances of action localization on ActivityNet dataset are reported on validation set.
The \textbf{THUMOS14} dataset includes 1,010 videos for validation and 1,574 videos for testing from 20 classes. Among all the videos, there are 220 and 212 videos with temporal annotations in validation and testing set, respectively.
The \textbf{Kinetics-600} is a large-scale action recognition dataset which consists of around 480K videos from 600 action categories. The 480K videos are divided into 390K, 30K, 60K for training, validation and test sets, respectively. Each video in the dataset is a 10-second clip of action moment annotated from raw YouTube video.

\subsection{Experimental Settings}

\textbf{Data Splits.} For each setting, our AherNet involves two datasets, untrimmed video set with temporal annotations and action moment set with only category labels. In the first setting, we split the classes of ActivityNet v1.3 into two parts according to the dataset taxonomy. The untrimmed video set (ANet-UN) contains 87 classes and the action moment set (ANet-AM) consists of the remaining 113 classes. We extract the foreground segments of training videos from 113 classes as the training data and take the original videos in the validation set from 113 classes as the validation data. In view that we aim to transfer action localization capability on the categories in ANet-UN to those in ANet-AM, this setting is named as ANet-UN$\rightarrow$ANet-AM. The second setting treats all the 220 validation videos in THUMOS14 (TH14) as untrimmed video set and the foreground segments of all the training videos in ActivityNet v1.3 as action moment set (ANet-FG). All the validation videos in ActivityNet v1.3 are exploited as the validation data. Similarly, we name this setting as TH14$\rightarrow$ANet-FG. In the third setting, we utilize ActivityNet v1.3 (ANet) and Kinetics-600 (K600) as untrimmed video set and action moment set, respectively. To verify action localization on 600 categories in Kinetics-600, we crawled at least 10 raw YouTube videos of action moments in validation set for each class. In total, the validation data contains 6,459 videos. This setting is namely ANet$\rightarrow$K600 for short.

\textbf{Implementations.} We utilize Pseudo-3D~\cite{Qiu:ICCV17} network as our 3D ConvNet for clip-level feature extraction.
The network input is a 16-frame clip and the sample rate of frames is set as $8$. The 2,048-way outputs from pool5 layer are extracted as clip-level features. During training, we choose three temporal scale ratios $\{r_d\}_{d=1}^{3}=[2^0, 2^{1/3}, 2^{2/3}]$ derived from~\cite{LinYi:ICCV17}. The parameter $\alpha_1$ and $\alpha_2$ are set as $1.0$ by cross validation. The threshold of NMS is set as 0.90. We implement our AherNet on Tensorflow~\cite{Abadi:TF} platform. In all the experiments, our networks are trained by utilizing adaptive moment estimation optimizer (Adam) \cite{Diederick:ICLR15}. The initial learning rate is set as $0.0001$, and decreased by $10\%$ after every $5k$ on first two data split settings and $15k$ on the final setting. The mini-batch size is $16$.

\textbf{Evaluation Metrics.} On all the three settings, we employ the mean average precision (mAP) values with IoU thresholds between $0.5$ and $0.95$ (inclusive) with a step size $0.05$ as the metric for comparison.

\begin{figure}
\begin{minipage}[!tb]{.45\linewidth}
\centering
       \scalebox{0.50}[0.50]{
              \begin{tabular}{{|l|c|c|}}
                     \hline
                     \multicolumn{1}{|c|}{\multirow{1}{*}{\text{Approach}}} & \multicolumn{1}{c|}{\text{{ANet-UN $\rightarrow$ ANet-AM}}} & \multicolumn{1}{c|}{\text{{TH14 $\rightarrow$ ANet-FG}}} \\ \hhline{*{3}{-}}
                     AherNet$^{0}$                   &       10.2          &     9.3             \\
                     AherNet$^{-}$                   &       12.8          &     10.4            \\ \hhline{*{3}{-}}
                     \text{AherNet,1-fc,none}        &       16.1          &     23.2            \\
                     \text{AherNet,1-fc,ReLU}        &       16.4          &     23.4            \\
                     \text{AherNet,1-fc,LeakyReLU}   &       16.5          &     23.5            \\
                     \text{AherNet,1-fc,ELU}         &       16.7          &     23.9            \\
                     \text{AherNet,2-fc,LeakyReLU}   &       16.9          &     24.2            \\
                     \text{AherNet,2-fc,ELU}         &    \textbf{17.2}    &    \textbf{24.3}    \\
                     \text{AherNet,3-fc,ELU}         &       16.8          &     24.1            \\ \hhline{*{3}{-}}
                     AherNet$^{*}$                   &       22.6          &     28.9            \\ \hhline{*{3}{-}}
              \end{tabular}

       }
       \captionof{table}{Exploration of different implementations of the weight transfer function in our AherNet. (fc means fully-connected layer).}
       \label{table4:1}
\end{minipage}
\begin{minipage}[!tb]{.55\linewidth}
       \centering
       \includegraphics[width=0.49\textwidth]{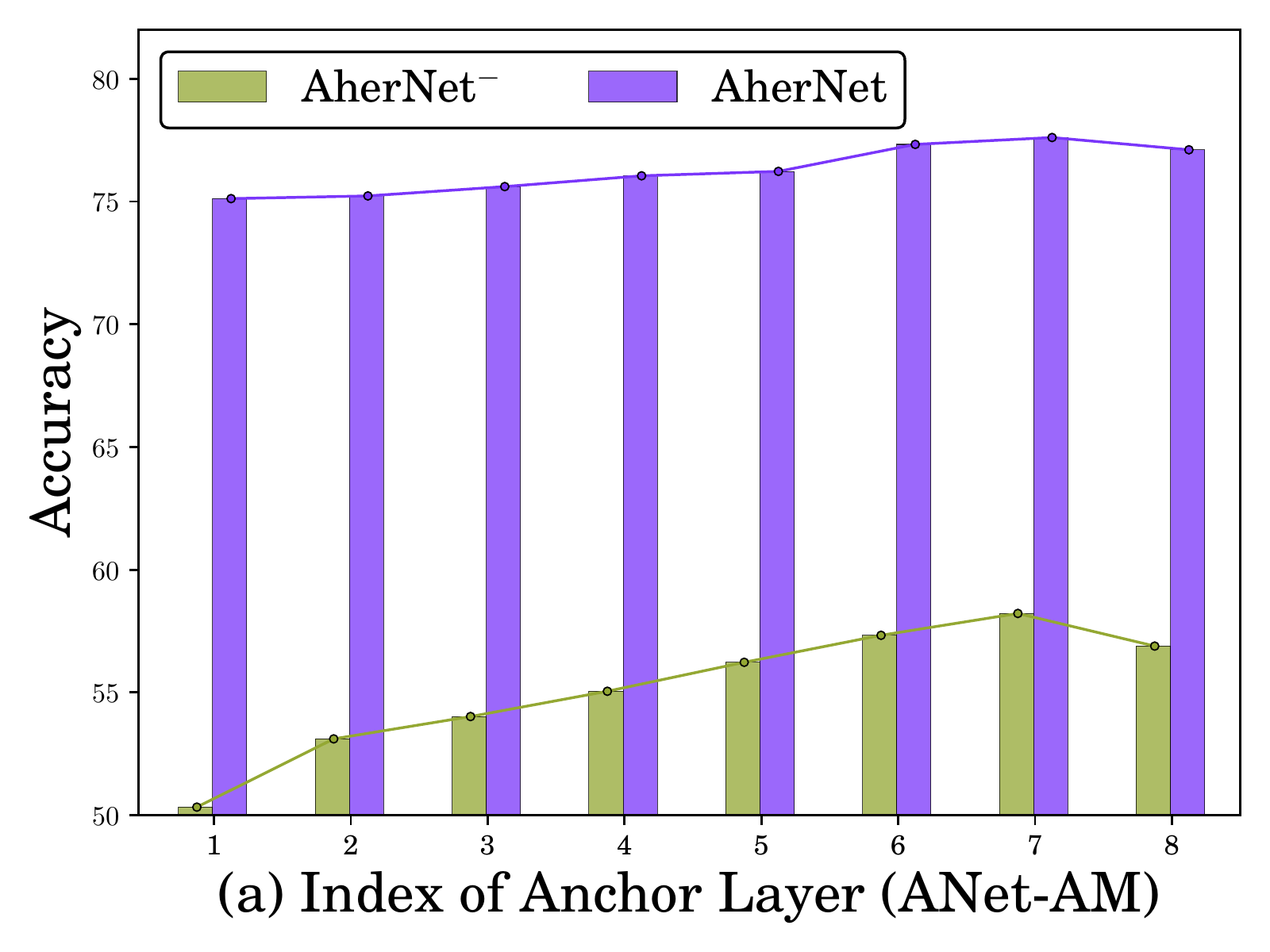}
       \includegraphics[width=0.49\textwidth]{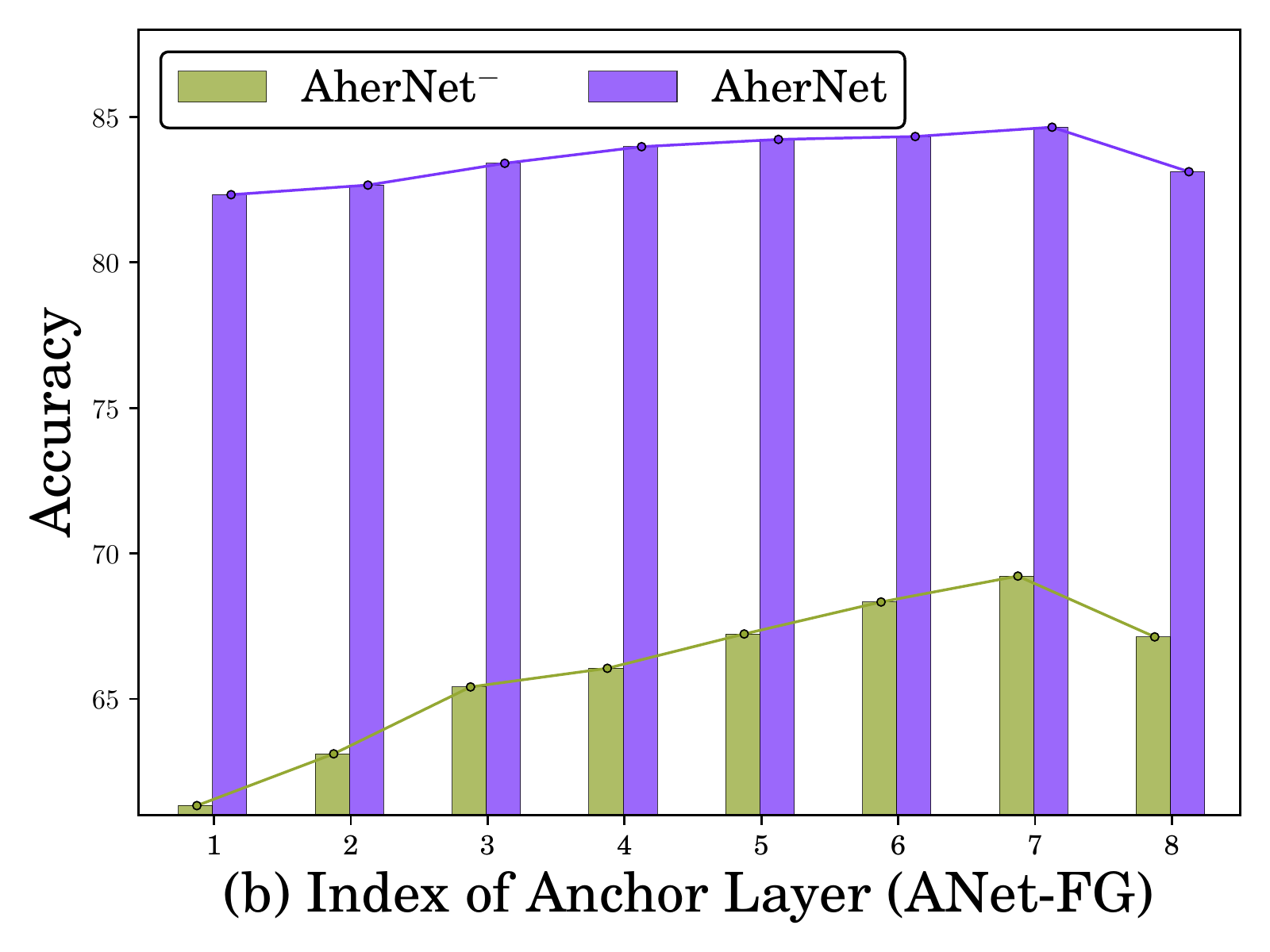}
       \caption{Action classification accuracy of AherNet$^-$ and AherNet in different anchor layer on (a) ANet-UN$\rightarrow$ANet-AM and (b) TH14$\rightarrow$ANet-FG.}
       \label{fig4:1}
\end{minipage}
\end{figure}

\subsection{Evaluation on Weight Transfer}

We first examine the module of weight transfer between classification and localization in our AherNet. We compare several implementations of the weight transfer function $\mathcal{T}$, e.g., different number of fully-connected layers plus various activation functions (ReLU, LeakyReLU \cite{Maas:ICML2013} and ELU \cite{Clevert:ICLR2016}), and three baseline approaches of AherNet$^0$, AherNet$^-$ and AherNet$^*$.
AherNet$^0$ is a purely classification-based model which learns a snippet-level classifier to predict the action score sequentially and splits action instances with multi-threshold strategy on the score sequence. As such, AherNet$^0$ is regarded as the lower bound. AherNet$^-$ deploys a ``proposal$+$classification'' scheme without weight transfer module. The action proposal model in AherNet$^-$ is learnt on untrimmed video set and directly performed on validation videos to output temporal action proposals. The classifier trained on action moment set is employed to predict the category of each action proposal.
AherNet$^*$ is an oracle run that exhaustively exploits the original videos of moment and trains a localization model in a fully-supervised manner. From this view, AherNet$^*$ is considered as the upper~bound.

Table \ref{table4:1} summarizes the average mAP performances over all IoU thresholds of different methods on the first two settings. AherNet with weight transfer function of two fully-connected layers plus ELU activation consistently exhibits better performance than other implementations across the two settings.
As expected, AherNet$^0$ performs worst since the method solely capitalizes on classification for localization problem without any knowledge of temporal action proposal. With the use of action proposal model learnt on untrimmed video set, AherNet$^-$ surpasses AherNet$^0$ by 2.6\% and 1.1\% on the settings of ANet-UN$\rightarrow$ANet-AM and TH14$\rightarrow$ANet-FG.
AherNet further boosts up the average mAP from 12.8\% and 10.4\% of AherNet$^-$ to 17.2\% and 24.3\%, respectively.
The results verify the merit of weight transfer in AherNet for bridging classification and localization, and scaling action localization to a large set of categories with only action moments.
In practice, AherNet has great potential to support localization for thousands of categories.
More importantly, when evaluating action localization model on the categories with full temporal annotation in the training, AherNet slightly outperforms AherNet$^*$, e.g., 25.4\% vs. 25.2\% and 27.7\% vs. 26.9\% on the actions in ANet-UN and TH14. This also demonstrates the advantage of leveraging action moments data in AherNet training to enhance action localization~model.

\begin{figure}
\begin{minipage}[!tb]{.40\linewidth}
\centering
       \scalebox{0.58}[0.58]{
              \begin{tabular}{{|l|c|c|c|c|}}
                     \hline
                     \multicolumn{1}{|c|}{\multirow{2}{*}{\text{Approach}}} & \multicolumn{2}{c|}{\text{ANet-UN $\rightarrow$ ANet-AM}} & \multicolumn{2}{c|}{\text{TH14 $\rightarrow$ ANet-FG}} \\ \hhline{*{1}{~}*{4}{-}}
                     \multicolumn{1}{|c|}{}             &~   \text{AUC} ~   & \text{mAP}          &~   \text{AUC}   ~ & \text{mAP}                \\ \hhline{*{5}{-}}
                     AherNet$^{0}$           &       41.8     &         10.2        &     11.2            &  9.3             \\
                     AherNet$^-$             &       52.6     &         12.8        &     16.4            &  10.4                \\ \hhline{*{5}{-}}
                     AherNet$_M$             &       53.5     &         13.2        &     49.7            &  17.3                \\ 
                     AherNet$_{A-}$          &       54.6     &         14.7        &     51.0            &  19.1                \\ 
                     AherNet                 & \textbf{58.3}  &   \textbf{17.2}     & \textbf{55.5}       &  \textbf{24.3}       \\ \hhline{*{5}{-}}
                     AherNet$^*$             &       61.1     &         22.6        &     63.4            &  28.9                \\ \hhline{*{5}{-}}
              \end{tabular}
       }
       \captionof{table}{The evaluations of localization modeling of AherNet.}
       \label{table4:2}
\end{minipage}
\begin{minipage}[!tb]{.60\linewidth}
\centering
        \includegraphics[width=0.45\textwidth]{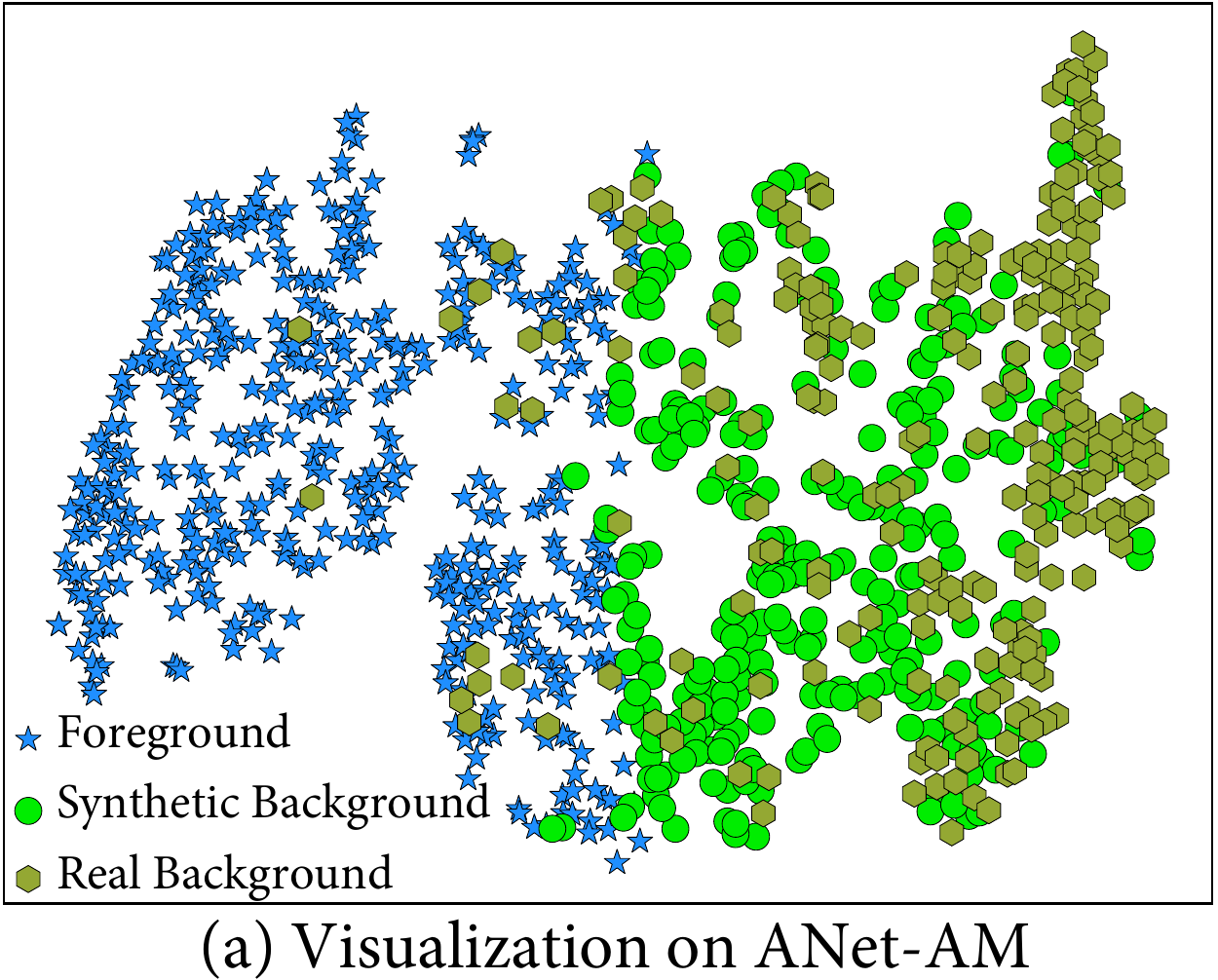}
        \includegraphics[width=0.45\textwidth]{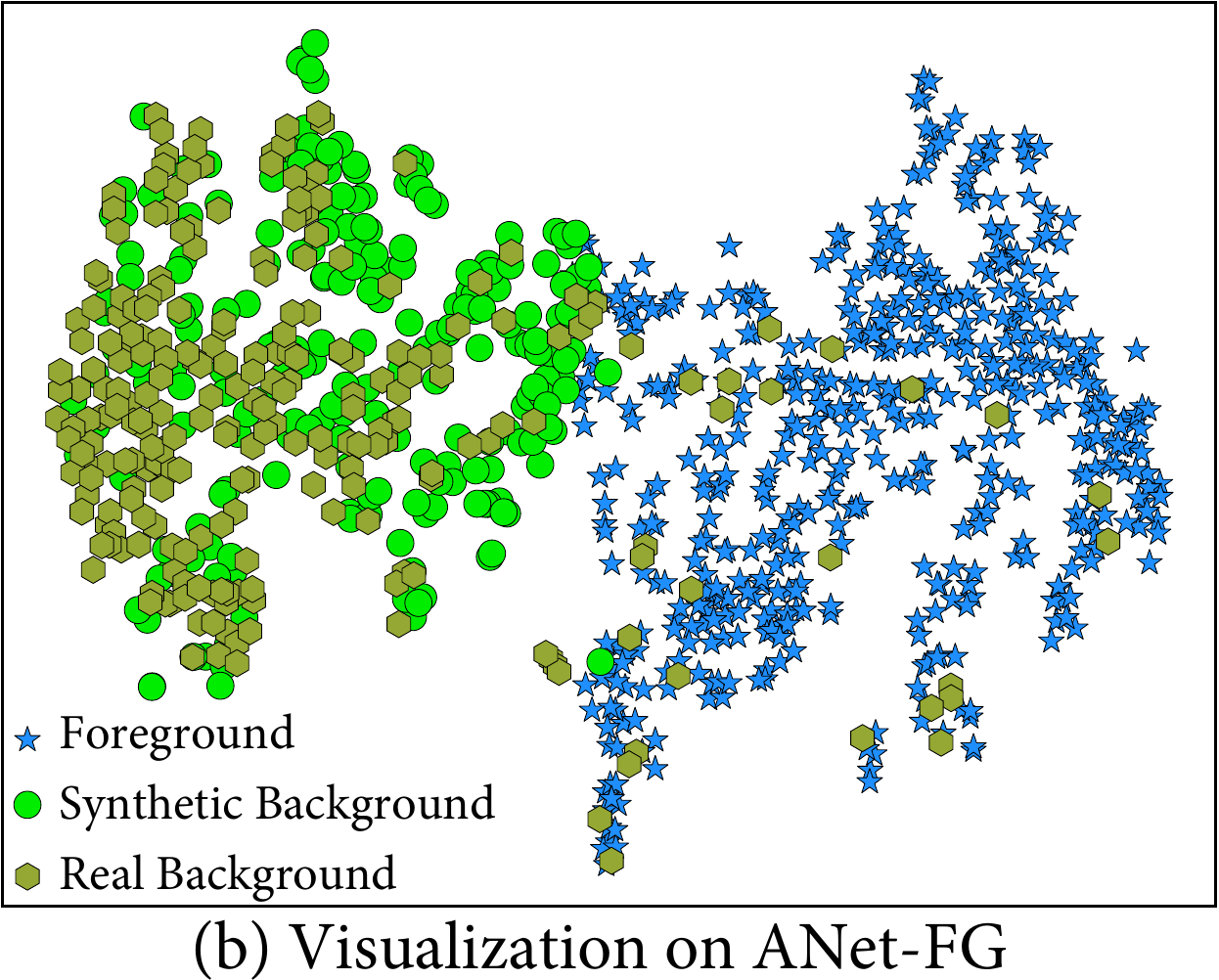}
        \caption{Feature visualization of AherNet: (a) ANet-UN$\rightarrow$ANet-AM~and (b)TH14$\rightarrow$ANet-FG.}
        \label{fig4:2}
\end{minipage}
\end{figure}

Figure \ref{fig4:1} further details the average classification accuracy over all proposals in each anchor layer.
Specifically, we feed the same proposals generated by AherNet into the classifier of AherNet$^-$ for accuracy computation on the same scale.
Because most proposals in ActivityNet range over about 40\% of the whole videos and such receptive field is nicely characterized by each anchor in the 7$^{th}$ layer, it is not a surprise that both AherNet and AherNet$^-$ achieve the highest accuracy on that layer.
Benefited from the capture of contexts in joint optimization with temporal action proposal, AherNet leads to better and more stable performances than AherNet$^-$.
The results again validate the weight transfer module.

\subsection{Evaluation on Localization Modeling}

Next, we study how localization modeling with context generation in AherNet influences the performances of both temporal action proposal and temporal action localization.
We design two additional runs of AherNet$_M$ and AherNet$_{A-}$ for comparison.
AherNet$_M$ capitalizes on only action moment set and directly learns an anchor-based action localization network by considering the starting/ending points of each moment as the time stamps of the action.
AherNet$_{A-}$ is a variant of AherNet by removing adversarial learning. The context generator is pre-trained on untrimmed video set through minimizing $L2$ loss between the converted background from foreground and the real background.

Table \ref{table4:2} shows the measure of area under Average Recall vs. Average Number of proposals per video curves (AUC) for action proposal and mAP performances for action localization.
Overall, AherNet$_M$ leads to a performance boost against AherNet$^-$ on both settings.
In particular, AherNet$_M$ improves the AUC value from 16.4\% to 49.7\% on TH14$\rightarrow$ANet-FG. Such results basically indicate that AherNet$_M$ is a practical choice for learning action localization directly on moment data.
AherNet$_{A-}$ is benefited from context generation for action moment set and the gain of mAP over AherNet$_M$ is 1.5\% and 1.8\%, respectively. Moreover, the upgrade of context generator from pre-training solely on untrimmed videos in AherNet$_{A-}$ to adversarial learning across the two video sets in AherNet contributes a mAP increase of 2.5\% and 5.2\%.

\begin{figure}
\begin{minipage}[!tb]{.40\linewidth}
\centering
       \scalebox{0.50}[0.50]{
        \begin{tabular}{{c|@{~~~}c@{~~~}c@{~~~}c@{~~~}|c}}
        \hline
        \multicolumn{5}{c}{\multirow{1}{*}{\textbf{ActivityNet v1.3, mAP}@$\alpha$}} \\ \hhline{*{5}{-}}
        \multicolumn{1}{c|@{~~~}}{Approach} & 0.5 & 0.75 & 0.95 & Average \\ \hline \hline
        \multicolumn{5}{c}{\multirow{1}{*}{\text{Fully-supervised Localization}}}                      \\ \hline
        \text{Wang~\emph{et al.}~\cite{Wang:anet16}}                &  45.11 & 4.11  & 0.05  & 16.41    \\
        \text{Singh~\emph{et al.}~\cite{Singh:CVPR16}}              &  26.01 & 15.22 & 2.61  & 14.62   \\
        \text{Singh~\emph{et al.}~\cite{Singh:ARXIV16}}             &  22.71 & 10.82 & 0.33  & 11.31   \\
        \text{CDC~\cite{Shou:CVPR17}}                               &  45.30 & 26.00 & 0.20  & 23.80   \\
        \text{TAG-D~\cite{Xiong:ARXIV17}}                           &  39.12 & 23.48 & 5.49  & 23.98   \\
        \text{Lin~\emph{et al.}~\cite{Lin:ARXIV17}}                 &  48.99 & 32.91 & 7.87  & 32.26   \\
        \text{BSN~\cite{Lin:ECCV18}}                                &  52.50 & 33.53 & 8.85  & 33.72   \\ \hline
        \multicolumn{5}{c}{\multirow{1}{*}{\text{Weakly-supervised Localization}}}                     \\ \hline
        \text{STPN~\cite{Nguyen:CVPR18}}                            &  29.30 & 16.90 & 2.60 & -        \\
        \text{Nguyen~\emph{et al.}~\cite{Nguyen:ICCV19}}            &  36.40 & 19.20 & 2.90 & -        \\ \hline
        \multicolumn{5}{c}{\multirow{1}{*}{\text{Partially-supervised Localization}}} \\ \hline
        \text{AherNet}         &  \textbf{40.33}  &\textbf{25.04} & \textbf{3.92}  &  \textbf{24.31}       \\ \hhline{*{5}{-}}
         \end{tabular}
       }
       \captionof{table}{Temporal action detection performances on ActivityNet v1.3, measured by mAP at different IoU thresholds $\alpha$.} 
       \label{table4:3}
\end{minipage}
\begin{minipage}[!tb]{.60\linewidth}
       \centering
       \includegraphics[width=0.492\textwidth]{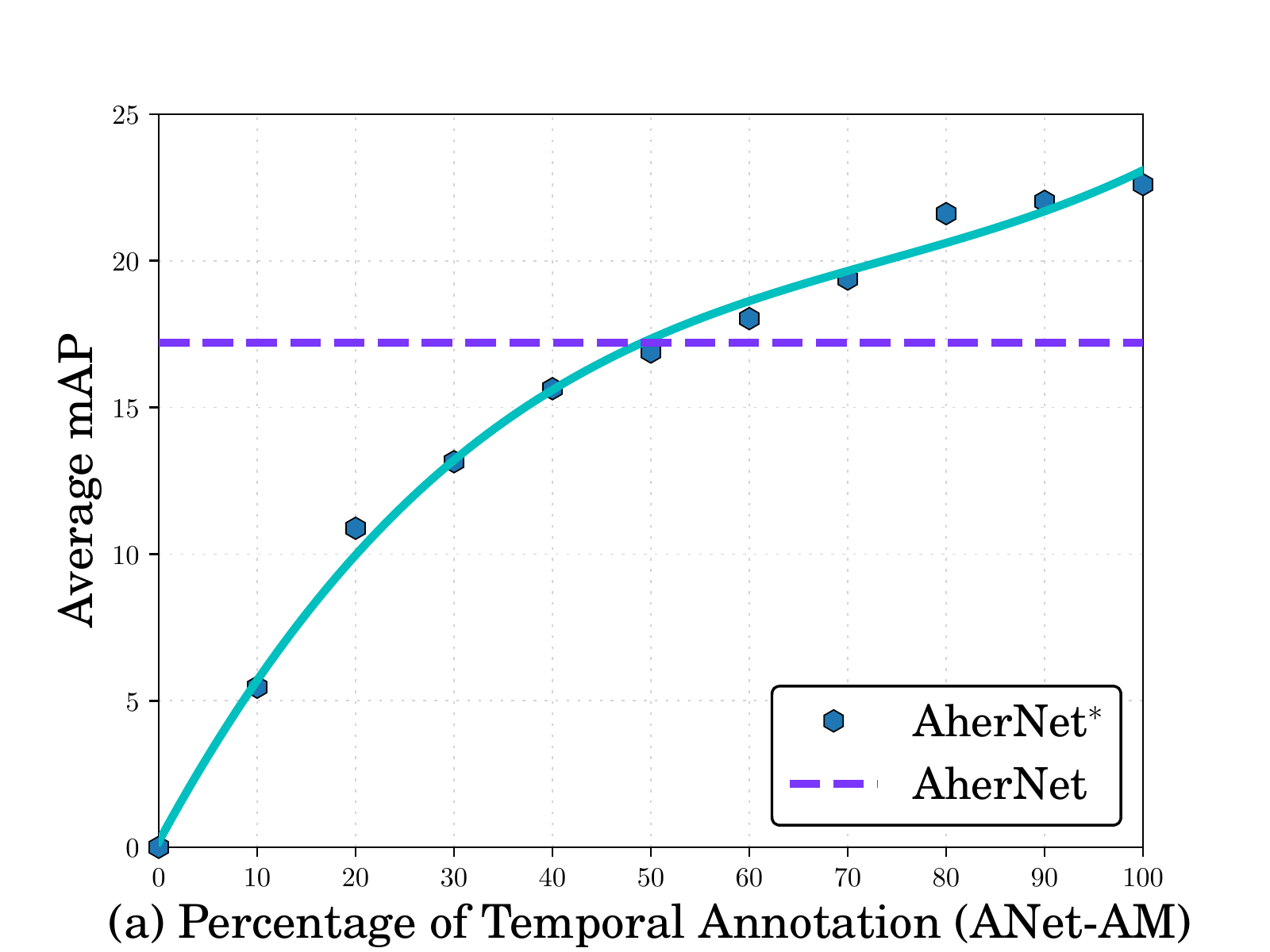}
       \includegraphics[width=0.492\textwidth]{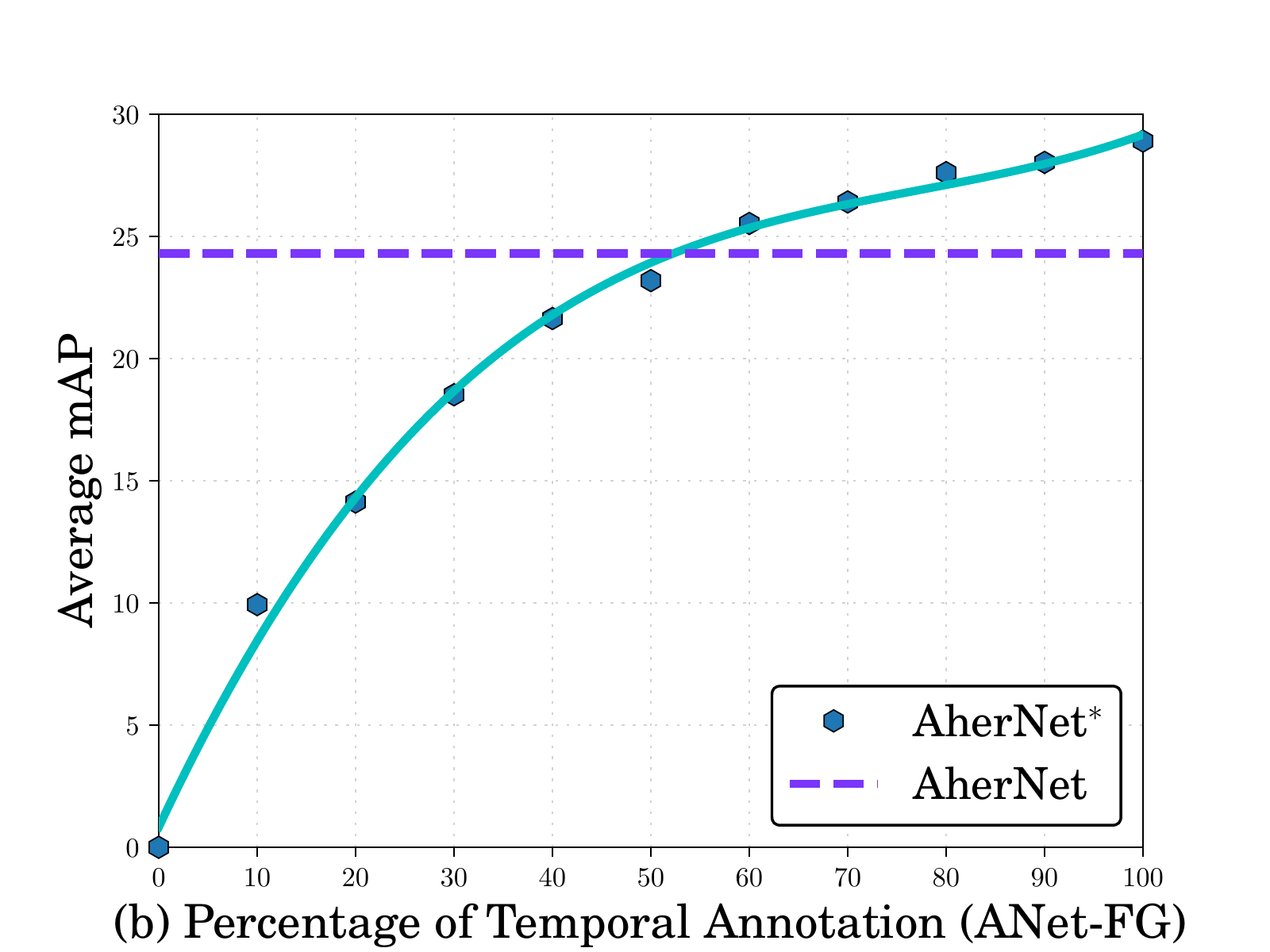}
       \caption{Average mAP comparisons of AherNet$^*$ learnt with different ratio of temporal annotation and AherNet, on (a) ANet-UN$\rightarrow$ANet-AM and (b) TH14$\rightarrow$ANet-FG.}
       \label{fig4:4}
\end{minipage}
\end{figure}

To examine the generated features of background, we further visualize the features of foreground, synthetic and real background for action moments by using t-SNE \cite{Maaten:JMLR8}. Specifically, we randomly select 500 anchors of foreground, synthetic and real background from 200 moments and the original videos in validation data, respectively. The first 256 principal components of the features of each anchor are extracted by PCA and projected into 2D space using t-SNE as shown in Figure \ref{fig4:2}. It is clear that the generated features of background by AherNet are indistinguishable from those of real background on both ANet-AM and ANet-FG sets, that confirms the effectiveness of context generation.

\subsection{Evaluation on Model Capacity of AherNet}
We discuss our AherNet with several state-of-the-art fully-supervised and weakly-supervised action localization methods. Table \ref{table4:3} lists the mAP performances under different IoU thresholds on ActivityNet v1.3 and such evaluation corresponds to the second setting of TH14$\rightarrow$ANet-FG for AherNet. The goal of weakly-supervised methods is to train action localization models for a set of categories which have untrimmed videos with only video-level labels. Instead, our AherNet enables the training of localization model for the categories of interest with action moments from these categories (e.g., ANet-FG) and temporal annotations from a small set of classes (e.g., TH14). Compared to the most recent advance \cite{Nguyen:ICCV19} in weakly-supervised localization, AherNet leads to a mAP boost of 3.9\% and 5.8\% under the IoU of 0.5 and 0.75, respectively.
AherNet is also comparable or even superior to several fully-supervised localization models, e.g., \cite{Shou:CVPR17} and \cite{Xiong:ARXIV17}, which rely on full temporal annotations for all the categories.
More importantly, the partially-supervised learning paradigm of our AherNet extends action localization to potentially thousands of categories in a more deployable~way.

To further quantitatively analyze the capability of AherNet, we compare AherNet with the fully-supervised version of AherNet$^*$ trained on different proportions of temporal annotations as shown in Figure \ref{fig4:4}. As expected, the average mAP performances of AherNet$^*$ constantly improve with respect to the increase of temporal annotations in training on both datasets. The results are desirable in the way that AherNet$^*$ starts to surplus the performance of AherNet till more than 50\% temporal annotations are leveraged.

\begin{figure*}[!tb]
       \centering\includegraphics[width=0.93\textwidth]{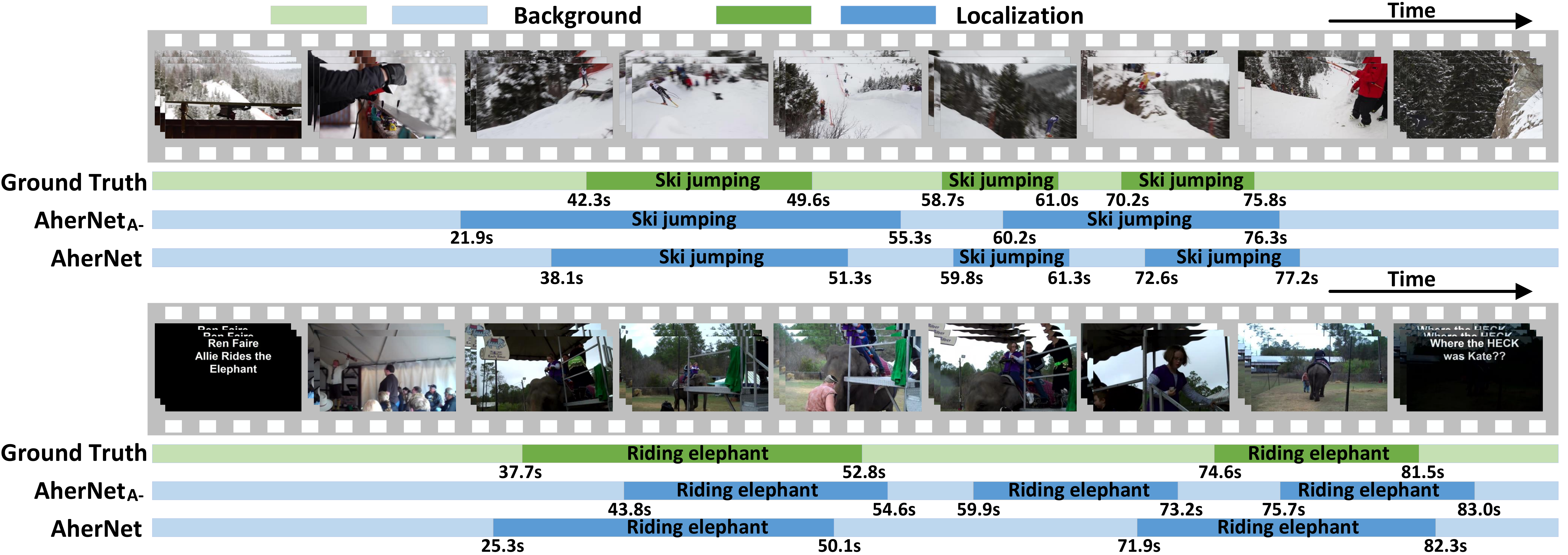}
       \caption{Example of two action localization results on Kinetics-600.}
\label{fig4:3}
\end{figure*}

\subsection{Large-Scale Action Localization of AherNet}
We finally take a step further to learn action localization model for 600 actions in Kinetics-600 dataset, which refers to the third setting of ANet$\rightarrow$K600.
Since the temporal annotations are not available for the validation videos of Kinetics-600, we collected the raw YouTube videos of action moments in our validation set and invited ten evaluators to label annotations.
Table \ref{table4:5} summarizes the mAP at different IoU thresholds on the setting of ANet$\rightarrow$K600. The performance trends are similar with those on the first two settings. AherNet boosts up the average mAP from 14.18\% to 24.43\%, indicating the impact of AherNet on the generalization of action localization for a large set of categories.
Figure \ref{fig4:3} showcases localization results of two videos from Kinetics-600, showing that AherNet nicely models the temporal dynamics and predicts accurate temporal boundaries.

\begin{table}[!tb]
       \setlength{\belowcaptionskip}{-1pt}
       \centering
       \caption{Performance comparisons of temporal action localization on Kinetics-600, measured by mAP at different IoU thresholds $\alpha$.}
       \scalebox{0.78}[0.78]{
        \begin{tabular}{{c|@{~~~~~~}c@{~~~~~~}c@{~~~~~~}c@{~~~~~~}|c}}
        \hline
        \multicolumn{5}{c}{\multirow{1}{*}{ANet$\rightarrow$K600, \textbf{Kinetics-600, mAP}@$\alpha$}} \\ \hhline{*{5}{-}}
        \multicolumn{1}{c|@{~~~~~~}}{Approach} & 0.5 & 0.75 & 0.95 & Average \\ \hline \hline
         AherNet$^0$            &     19.26        &    16.72      &       0.88      &   14.18            \\
         AherNet$^-$            &     21.76        &    17.85      &       1.71      &   15.96            \\
         AherNet$_M$            &     28.98        &    20.71      &       2.95      &   19.05            \\
         AherNet$_{A-}$         &     32.71        &    23.04      &       5.08      &   21.68            \\ \hhline{*{5}{-}}
         AherNet                &  \textbf{36.19}  &\textbf{26.96} & \textbf{6.55}  &  \textbf{24.43}    \\ \hhline{*{5}{-}}
         \end{tabular}
       }
    \label{table4:5}
\end{table}

\section{Conclusions}
We have presented Action Herald Networks (AherNet) which scale action localization to a large set of categories. Particularly, we study the problem from a new learning paradigm of training localization model with only trimmed action moments from the large set of categories plus temporal annotations on untrimmed videos from a small set of action classes. To materialize our idea, we have devised an one-stage action localization framework which consists of two key modules: weight transfer between classification and localization, and localization modeling on action moments. The former extracts foreground segments from untrimmed videos as action moments, and learns a weight transfer function between foreground segment classification and temporal action classification in localization. The latter simulates action localization on action moments data by hallucinating the background features of an action moment via adversarial learning.
Experiments conducted on two settings, i.e., across the splits of ActivityNet v1.3 and from THUMOS14 to ActivityNet v1.3, validate our proposal.
More remarkably, we build a large-scale localization model for 600 categories in Kinetics-600.

\textbf{Acknowledgements.} This work is partially supported by Beijing Academy of Artificial Intelligence (BAAI) and the National Key R\&D Program of China under contract No. 2017YFB1002203.

%
%
\bibliographystyle{splncs04}
\bibliography{egbib}

\begin{thebibliography}{10}
\providecommand{\url}[1]{\texttt{#1}}
\providecommand{\urlprefix}{URL }
\providecommand{\doi}[1]{https://doi.org/#1}

\bibitem{Abadi:TF}
Abadi, M., Barham, P., Chen, J., Chen, Z., Davis, A., Dean, J., Devin, M.,
  Ghemawat, S., Irving, G., Isard, M., Kudlur, M., Levenberg, J., Monga, R.,
  Moore, S., Murray, D.G., Steiner, B., Tucker, P., Vasudevan, V., Warden, P.,
  Wicke, M., Yu, Y., Zheng, X.: {TensorFlow: A System for Large-scale Machine
  Learning}. In: OSDI (2016)

\bibitem{Buch:BMVC17}
Buch, S., Escorcia, V., Ghanem, B., Fei-Fei, L., Niebles, J.C.: {End-to-End,
  Single-Stream Temporal Action Detection in Untrimmed Videos}. In: BMVC (2017)

\bibitem{Buch:CVPR17}
Buch, S., Escorcia, V., Shen, C., Ghanem, B., Niebles, J.C.: {SST:
  Single-Stream Temporal Action Proposals}. In: CVPR (2017)

\bibitem{Chao:CVPR18}
Chao, Y.W., Vijayanarasimhan, S., Seybold, B., Ross, D.A., Deng, J.,
  Sukthankar, R.: {Rethinking the Faster R-CNN Architecture for Temporal Action
  Localization}. In: CVPR (2018)

\bibitem{Clevert:ICLR2016}
Clevert, D.A., Unterthiner, T., Hochreiter, S.: {Fast and Accurate Deep Network
  Learning by Exponential Linear Units (ELUs)}. In: ICLR (2016)

\bibitem{Escorcia:ECCV16}
Escorcia, V., Heilbron, F.C., Niebles, J.C., Ghanem, B.: {DAPs: Deep Action
  Proposals for Action Understanding}. In: ECCV (2016)

\bibitem{Gaidon:PAMI13}
Gaidon, A., Harchaoui, Z., Schmid, C.: {Temporal Localization of Actions with
  Actoms}. IEEE Trans. on PAMI  \textbf{35}(11),  2782--2795 (2013)

\bibitem{Ganin:ICML15}
Ganin, Y., Lempitsky, V.: {Unsupervised Domain Adaptation by Backpropagation}.
  In: ICML (2015)

\bibitem{Gao:ECCV18}
Gao, J., Chen, K., Nevatia, R.: {CTAP: Complementary Temporal Action Proposal
  Generation}. In: ECCV (2018)

\bibitem{Gao:ICCV17}
Gao, J., Yang, Z., Sun, C., Chen, K., Nevatia, R.: {TURN TAP: Temporal Unit
  Regression Network for Temporal Action Proposals}. In: ICCV (2017)

\bibitem{Geest:ECCV16}
Geest, R.D., Gavves, E., Ghodrati, A., Li, Z., Snoek, C., Tuytelaars, T.:
  {Online Action Detection}. In: ECCV (2016)

\bibitem{Kinetics:600}
Ghanem, B., Niebles, J.C., Snoek, C., Heilbron, F.C., Alwassel, H., Escorcia,
  V., Krishna, R., Buch, S., Dao, C.D.: {The ActivityNet Large-Scale Activity
  Recognition Challenge 2018 Summary}. arXiv preprint arXiv:1808.03766  (2018)

\bibitem{Girshick:ICCV15}
Girshick, R.: {Fast R-CNN}. In: ICCV (2015)

\bibitem{Goodfellow:NIPS14}
Goodfellow, I.J., Pouget-Abadie, J., Mirza, M., Xu, B., Warde-Farley, D.,
  Ozair, S., Courville, A., Bengio, Y.: {Generative Adversarial Nets}. In: NIPS
  (2014)

\bibitem{He:ICCV17}
He, K., Gkioxari, G., Dollar, P., Girshick, R.: {Mask R-CNN}. In: ICCV (2017)

\bibitem{Heilbron:CVPR17}
Heilbron, F.C., Barrios, W., Escorica, V., Ghanem, B.: {SCC: Semantic Context
  Cascade for Efficient Action Detection}. In: CVPR (2017)

\bibitem{ActivityNet}
Heilbron, F.C., Escorcia, V., Ghanem, B., Niebles, J.C.: {ActivityNet: A
  Large-Scale Video Benchmark for Human Activity Understanding}. In: CVPR
  (2015)

\bibitem{Hoffman:NIPS14}
Hoffman, J., Guadarrama, S., Tzeng, E., Hu, R., Donahue, J.: {LSDA: Large Scale
  Detection through Adaptation}. In: NIPS (2014)

\bibitem{Hu:CVPR18}
Hu, R., Dollar, P., He, K., Darell, T., Girshick, R.: {Learning to Segment
  Every Thing}. In: CVPR (2018)

\bibitem{Isola:CVPR17}
Isola, P., Zhu, J.Y., Zhou, T., Efros, A.A.: {Image-to-Image Translation with
  Conditional Adversarial Networks}. In: CVPR (2017)

\bibitem{Thumos}
Jiang, Y.G., Liu, J., R.Zamir, A., Toderici, G.: {THUMOS challenge: Action
  recognition with a large number of classes}.
  \url{http://crcv.ucf.edu/THUMOS14} (2014)

\bibitem{Diederick:ICLR15}
Kingma, D.P., Ba, J.: {Adam: A Method for Stochastic Optimization}. In: ICLR
  (2015)

\bibitem{Kuen:ICCV19}
Kuen, J., Perazzi, F., Lin, Z., Zhang, J., Tan, Y.P.: {Scaling Object Detection
  by Transferring Classification Weights}. In: ICCV (2019)

\bibitem{Lea:CVPR17}
Lea, C., Michael D.~Flynn, R.V., Reiter, A., Hager, G.D.: {Temporal
  Convolutional Netowrk for Action Segmentation and Detection}. In: CVPR (2017)

\bibitem{Dong:ECCV18}
Li, D., Qiu, Z., Dai, Q., Yao, T., Mei, T.: {Recurrent Tubelet Proposal and
  Recognition Networks for Action Detection}. In: ECCV (2018)

\bibitem{Dong:MM19}
Li, D., Yao, T., Qiu, Z., Li, H., Mei, T.: {Long Short-Term Relation Networks
  for Video Action Detection}. In: ACM MM (2019)

\bibitem{Lin:MM17}
Lin, T., Zhao, X., Shou, Z.: {Single Shot Temporal Action Detection}. In: ACM
  MM (2017)

\bibitem{Lin:ARXIV17}
Lin, T., Zhao, X., Shou, Z.: {Temporal convolution based action proposal:
  Submission to activitynet 2017}. arXiv preprint arXiv:1707.06750  (2017)

\bibitem{Lin:ECCV18}
Lin, T., Zhao, X., Su, H., Wang, C., Yang, M.: {BSN: Boundary Sensitive Network
  for Temporal Action Proposal Generation}. In: ECCV (2018)

\bibitem{LinYi:ICCV17}
Lin, T.Y., Goyal, P., Girshick, R., He, K., Dollar, P.: {Focal Loss for Dense
  Object Detection}. In: ICCV (2017)

\bibitem{Liu:CVPR19}
Liu, D., Jiang, T., Wang, Y.: {Completeness Modeling and Context Separation for
  Weakly Supervised Temporal Action Localization}. In: CVPR (2019)

\bibitem{Long:CVPR19}
Long, F., Yao, T., Qiu, Z., Tian, X., Luo, J., Mei, T.: {Gaussian Temporal
  Awareness Networks for Action Localization}. In: CVPR (2019)

\bibitem{Long:TMM20}
Long, F., Yao, T., Qiu, Z., Tian, X., Mei, T., Luo, J.: {Coarse-to-Fine
  Localization of Temporal Action Proposals}. IEEE Trans. on Multimedia
  \textbf{22}(6),  1577 -- 1590 (2020)

\bibitem{Lu:TMM14}
Lu, S., Wang, Z., Mei, T., Guan, G., Feng, D.D.: {A Bag-of-Importance Model
  With Locality-Constrained Coding Based Feature Learning for Video
  Summarization}. IEEE Trans. on Multimedia  \textbf{16}(6),  1497 -- 1509
  (2014)

\bibitem{Maas:ICML2013}
Maas, A.L., Hannun, A.Y., Ng, A.Y.: {Rectifier Nonlinearities Improve Neural
  Network Acoustic Models}. In: ICML (2013)

\bibitem{Maaten:JMLR8}
van~der Maaten, L., Hinton, G.: {Visualizing Data using t-SNE}. JMLR  (2008)

\bibitem{Nguyen:CVPR18}
Nguyen, P., Liu, T., Prasad, G., Han, B.: {Weakly Supervised Action
  Localization by Sparse Temporal Pooling Network}. In: CVPR (2018)

\bibitem{Nguyen:ICCV19}
Nguyen, P.X., Ramanan, D., Fowlkes, C.C.: {Weakly-supervised Action
  Localization with Background Modeling}. In: ICCV (2019)

\bibitem{Oneata:ICCV13}
Oneata, D., Verbeek, J., Schmid, C.: {Action and Event Recognition with Fisher
  Vectors on a Compact Feature Set}. In: ICCV (2013)

\bibitem{Qiu:ICCV17}
Qiu, Z., Yao, T., Mei, T.: {Learning Spatio-Temporal Representation with
  Pseudo-3D Residual Networks}. In: ICCV (2017)

\bibitem{Qiu:CVPR19}
Qiu, Z., Yao, T., Ngo, C.W., Tian, X., Mei, T.: {Learning Spatio-Temporal
  Representation with Local and Global Diffusion}. In: CVPR (2019)

\bibitem{Shi:CVPR20}
Shi, B., Dai, Q., Mu, Y., Wang, J.: {Weakly-Supervised Action Localization by
  Generative Attention Modeling}. In: CVPR (2020)

\bibitem{Shou:CVPR17}
Shou, Z., Chan, J., Zareian, A., Miyazawa, K., Chang, S.F.: {CDC:
  Convolutional-De-Convolutional Network for Precise Temporal Action
  Localization in Untrimmed Videos}. In: CVPR (2017)

\bibitem{Shou:ECCV18}
Shou, Z., Gao, H., Zhang, L., Miyazawa, K., Chang, S.F.: {AutoLoc:
  Weakly-supervised Temporal Action Localization in Untrimmed Videos}. In: ECCV
  (2018)

\bibitem{Shou:ECCV18B}
Shou, Z., Pan, J., Chan, J., Miyazawa, K., Mansour, H., Vetro, A., i~Nieto,
  X.G., Chang, S.F.: {Online Detection of Action Start in Untrimmed, Streaming
  Videos}. In: ECCV (2018)

\bibitem{Shou:CVPR16}
Shou, Z., Wang, D., Chang, S.F.: {Temporal Action Localization in Untrimmed
  Videos via Multi-stage CNNs}. In: CVPR (2016)

\bibitem{Singh:CVPR16}
Singh, B., Marks, T.K., Jones, M., Tuzel, O., Shao, M.: {A Multi-Stream
  Bi-Directional Recurrent Neural Network for Fine-Grained Action Detection}.
  In: CVPR (2016)

\bibitem{Singh:ARXIV16}
Singh, G., Cuzzolin, F.: {Untrimmed Video Classification for Activity
  Detection: submission to ActivityNet Challenge}. arXiv preprint
  arXiv:1607.01979  (2016)

\bibitem{Tang:ICCV13}
Tang, K., Yao, B., Fei-Fei, L., Koller, D.: {Combining the Right Features for
  Complex Event Recognition}. In: ICCV (2013)

\bibitem{Tang:CVPR16}
Tang, Y., Wang, J., Gao, B., Dellandrea, E., Gaizauskas, R., Chen, L.: {Large
  Scale Semi-supervised Object Detection using Visual and Semantic Knowledge
  Transfer}. In: CVPR (2016)

\bibitem{Tzeng:CVPR17}
Tzeng, E., Hoffman, J., Saenko, K., Darrell, T.: {Adversarial Discriminative
  Domain Adaption}. In: CVPR (2017)

\bibitem{Wang:CVPR17}
Wang, L., Xiong, Y., Lin, D., Gool, L.V.: {UntrimmedNets for Weakly Supervised
  Action Recognition and Detection}. In: CVPR (2017)

\bibitem{Wang:anet16}
Wang, R., Tao, D.: {UTS} at activitynet 2016. In: CVPR ActivityNet Challenge
  Workshop (2016)

\bibitem{Xiong:ARXIV17}
Xiong, Y., Zhao, Y., Wang, L., Lin, D., Tang, X.: {A Pursuit of Temporal
  Accuracy in General Activity Detection}. arXiv preprint arXiv:1703.02716
  (2017)

\bibitem{Xu:ICCV17}
Xu, H., Das, A., Saenko, K.: {R-C3D: Region Convolutional 3D Network for
  Temporal Activity Detection}. In: ICCV (2017)

\bibitem{Yeung:CVPR16}
Yeung, S., Russakovsky, O., Mori, G., Fei-Fei, L.: {End-to-end Learning of
  Action Detection from Frame Glimpses in Videos}. In: CVPR (2016)

\bibitem{Yuan:CVPR16}
Yuan, J., Ni, B., Yang, X., A.Kassim, A.: {Temporal Action Localization With
  Pyramid of Score Distribution Features}. In: CVPR (2016)

\bibitem{Zeng:ICCV19}
Zeng, R., Huang, W., Tan, M., Rong, Y., Zhao, P., Huang, J., Gan, C.: {Graph
  Convolutional Networks for Temporal Action Localization}. In: ICCV (2019)

\bibitem{Xiong:ICCV17}
Zhao, Y., Xiong, Y., Wang, L., Wu, Z., Tang, X., Lin, D.: {Temporal Action
  Detection with Structured Segment Networks}. In: ICCV (2017)

\end{thebibliography}
\end{document}